\documentclass[runningheads]{llncs}
\usepackage{graphicx}
\usepackage{tikz}
\usepackage{comment}
\usepackage{amsmath,amssymb} % define this before the line numbering.
\usepackage{color}
\usepackage{siunitx}

\begin{document}
% \linenumbers
\pagestyle{headings}
\mainmatter
\def\ECCVSubNumber{100}  % Insert your submission number here

\title{Adversarial Model for Rotated Indoor Scenes Planning} % Replace with your title

%\begin{comment}
%\titlerunning{ECCV-20 submission ID \ECCVSubNumber} 
%\authorrunning{ECCV-20 submission ID \ECCVSubNumber} 
%\author{Anonymous ECCV submission}
%\institute{Paper ID \ECCVSubNumber}
%\end{comment}
%******************

%\begin{comment}
%\author{Xinhan Di\inst{1}\orcidID{0000-1111-2222-3333} \and
%Second Author\inst{2,3}\orcidID{1111-2222-3333-4444} \and
%Third Author\inst{3}\orcidID{2222--3333-4444-5555}}
%\institute{Princeton University, Princeton NJ 08544, USA \and
%Springer Heidelberg, Tiergartenstr. 17, 69121 Heidelberg, Germany
%\end{comment}
\author{Xinhan Di\inst{1}\and Pengqian Yu \inst{2} \and Hong Zhu \inst{1} \and Lei Cai  \inst{1} \and Qiuyan Sheng \inst{1}\and Changyu Sun \inst{1}}
\authorrunning{Xin.Di et al.}
\institute{Technique Center Ihome Corporation, Nanjing, China
\email{\{deepearthgo,jszh0825,caileitx1990\}@gmail.com,\\shenqiuyan123@foxmail.com,744370610@qq.com}\\
\and
IBM Research, Singapore\\
\email{peng.qian.yu@ibm.com}}
%******************
\maketitle
\begin{abstract}
In this paper, we propose an adversarial model for producing furniture layout for interior scene synthesis when the interior room is rotated. The proposed model combines a conditional adversarial network, a rotation module, a mode module, and a rotation discriminator module. As compared with the prior work on scene synthesis, our proposed three modules enhance the ability of auto-layout generation and reduce the mode collapse during the rotation of the interior room. We conduct our experiments on a proposed real-world interior layout dataset that contains $14400$ designs from the professional designers. Our numerical results demonstrate that the proposed model yields higher-quality layouts for four types of rooms, including the bedroom, the bathroom, the study room, and the tatami room.     
%10 figures 4 tables = 4.5 Pages%
\end{abstract}

\section{Introduction}
People spend lots of time indoors such as the bedroom, living room, office, gym and so on. Function, beauty, cost and comfort are keys for the redecoration of indoor scenes. Proprietor prefers demonstration of the layout of indoor scenes in several minutes nowadays. Therefore, online virtual interior tools become  useful to help people design indoor spaces. These tools are faster, cheaper and more flexible than real redecoration in the real-world scenes. This fast demonstration is often based on auto layout of in-door furniture and a good graphics engine. Machine learning researchers take use of virtual tools to train data-hungry models for the auto layout \cite{Dai_2018_CVPR,Gordon_2018_CVPR}. The models reduce the time of layout of furniture from hours to minutes and support the fast demonstration. 

Generative models of indoor scenes are valuable for the auto layout of the furniture. This problem of indoor scenes synthesis are studied since the last decade. One family of the approach is object-oritented which the objects in the space are represented explicitly \cite{10.1145/2366145.2366154,10.1145/3303766,Qi_2018_CVPR,10.1145/3303766}. The other family of models is space-oriented which space is treated as a first-class entity and each point in space is occupied through the modeling \cite{10.1145/3197517.3201362}.

\begin{figure}
\centering
\includegraphics[height=7.5cm]{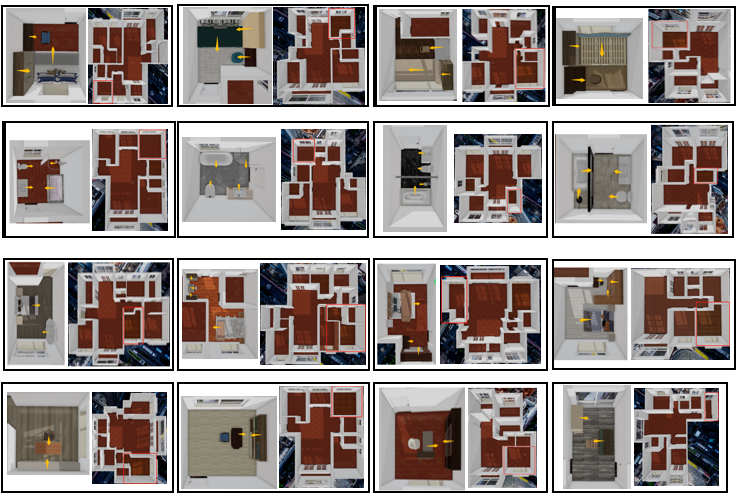}
\caption{Rooms are oriented towards different directions in the real word. Rooms examples represented in different rows are tatami, bathroom,  bedroom, and study room. In each example, a house floor plan is represented at the right side, and the room with layout is represented at the left side. These rooms are oriented towards different directions.}
\label{fig1}
\end{figure}

Deep generative modes are used for efficient generation of indoor scenes for auto-layout recently. These deep models further reduce the time from minutes to seconds. The variety of the generative layout is also increased. The deep generative models directly produces the layout of the furniture given an empty room. However, in the real world, the direction of a room is diverse in the real world. The south, north or northwest directions are equally possible. The layout for the real indoor scenes are required to meet with different directions as illustrated in Figure \ref{fig1}. 

Motivated by the above mentioned challenge, we propose an adversarial generative model for rotated indoor scenes in this paper. The model yields a design of layout of furniture when indoor scenes are rotated. In particular, this proposed adversarial model consists of several modules including rotation modules, two mode modules and double discriminators. The rotation modules are applied to the hidden layer of the generative models, and the mode modules are applied to the generative output and the ground truth. In addition, the double discriminators are applied to deal with the rotations in the indoor scenes.

This paper is organized as follows: the related work is discussed in the Section 2. Section 3 introduces the problem formulation. The methods of the proposed adversarial model are in the Section 4. The experiments and comparisons with baseline generative models can be found in Section 5. The paper is concluded with discussions in Section 6.

\section{Related Work}
Our work is related to data-hungry methods for synthesizing indoor scenes through the layout of furniture unconditionally or partially conditionally. 

\subsection{Structured data representation}
Representation of scenes as a graph is an elegant methodology since the layout of furniture for indoor scenes are highly structured. In the graph, semantic relationships are encoded as edges and objects are encoded as nodes. A small dataset of annotated scene hierarchies is learned as a grammar for the prediction of hierarchical indoor scenes \cite{10.1145/3197517.3201362}. Then, the generation of scene graphs from images is applied, including using scene graph for image retrieval \cite{Johnson_2015_CVPR} and generation of 2D images from an input scene graph \cite{Johnson_2018_CVPR}. However, the use of this family of structure representation is limited to a small dataset. In addition, it is not practical for the auto layout of furniture in the real world. 

\subsection{Indoor scene synthesis}
Early work in the scene modeling implemented kernels and graph walks to retrieve objects from a database \cite{Choi_2013_CVPR,Dasgupta_2016_CVPR}. The graphical models are employed to model the compatibility between furniture and input sketches of scenes \cite{10.1145/2461912.2461968}. However, these early methods are mostly limited by the scenes size. It is therefore hard to produce good-quality layout for large scene size. With the availability of large scene datasets including SUNCG \cite{Song_2017_CVPR}, more sophisticated learning methods are proposed as we review them below.

\subsection{Image CNN networks}
An image-based CNN network is proposed to encoded top-down views of input scenes, and then the encoded scenes are decoded for the prediction of object category and location \cite{10.1145/3197517.3201362}. A variational auto-encoder is applied to the generation of scenes with representation of a matrix. In the matrix, each column is represented as an object with location and geometry attributes \cite{10.1145/3381866}. A semantically-enriched image-based representation is learned from the top-down views of the indoor scenes, and convolutional object placement priors is trained \cite{10.1145/3197517.3201362}. However, this family of image CNN networks can not apply to the situation where rooms in the real world are located towards a variety of directions.

\subsection{Graph generative networks}
As a significant number of methods has been proposed to model graphs as networks \cite{DBLP:journals/corr/abs-1709-05584,4700287}, the family for the representation of indoor scenes in the form of tree-structured scene graphs are studied. For example, Grains \cite{10.1145/3303766} consists of a recursive auto-encoder network for the graph generation and it is targeted to produce different relationships including surrounding and supporting. Similarly, a graph neural network is proposed for scene synthesis. The edges is represented as spatial and semantic relationships of objects \cite{10.1145/3197517.3201362} in a dense graph. Both relationship graphs and  instantiation are generated for the design of indoor scenes. The relationship graph helps to find symbolical objects and the high-lever pattern \cite{10.1145/3306346.3322941}.

\subsection{CNN generative networks}
Layout of indoor scenes is also explored as the problem of generation of layout. Geometric relations of different types of 2D elements of indoor scenes are modeled through synthesis of layouts. This synthesis is trained through an adversarial network with self-attention modules \cite{DBLP:journals/corr/abs-1901-06767}. A variational autoencoder is proposed for the generation of stochastic scene layouts with prior of a label for each scene \cite{Jyothi_2019_ICCV}. However, the generation of layout is limited to single direction of the indoor scenes, while the real scenes can have various directions. 

\section{Problem Formulation}
\begin{figure}
\centering
\includegraphics[height=4.5cm]{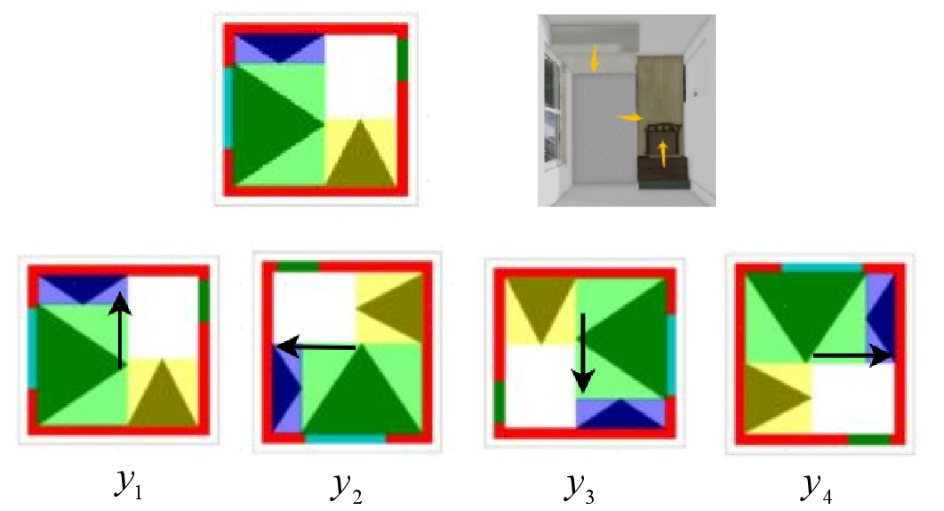}
\caption{Rooms are located towards four different directions with the layout of furniture. The positions, directions of each furniture, wall, door and window are represented.}
\label{fig2}
\end{figure}
We let a set of indoor scenes ${(x_{1},y_{1}, \theta_{1}),\dots,(x_{N},y_{N},\theta_{N})}$ where $N$ is the number of the scenes, and $x_{i}$ is an empty indoor scene with basic elements including walls, doors and windows. $y_{i}$ is the corresponding layout of the furniture for $x_{i}$. Each $y_{i}$ contains the elements ${p_{j},s_{j},d_{j}}$: $p_{j}$ is the position of the $jth$ element; $s_{j}$ is the size of the $jth$ element; and $d_{j}$ is the direction of $jth$ element. Each element represents a furniture in an indoor scene $i$. We use $\theta_{i} $ to indicate the direction of the indoor scene $i$. Figure \ref{fig2} illustrates four instances of $y_{i}$  that represents the direction of an indoor scene including the position, size, directions of each furniture, walls, doors and windows in that scene.

We define a model $M$ such that $y_{pre} = M(x_{in},\theta_{in})$, and given an empty room $x_{in}$ with walls, windows and doors, and the direction of the room $\theta_{in}$, the model $M$ produces layout $y_{pre}$ including the position, size, and direction of each furniture.

\begin{figure}
\centering
\includegraphics[height=8.5cm]{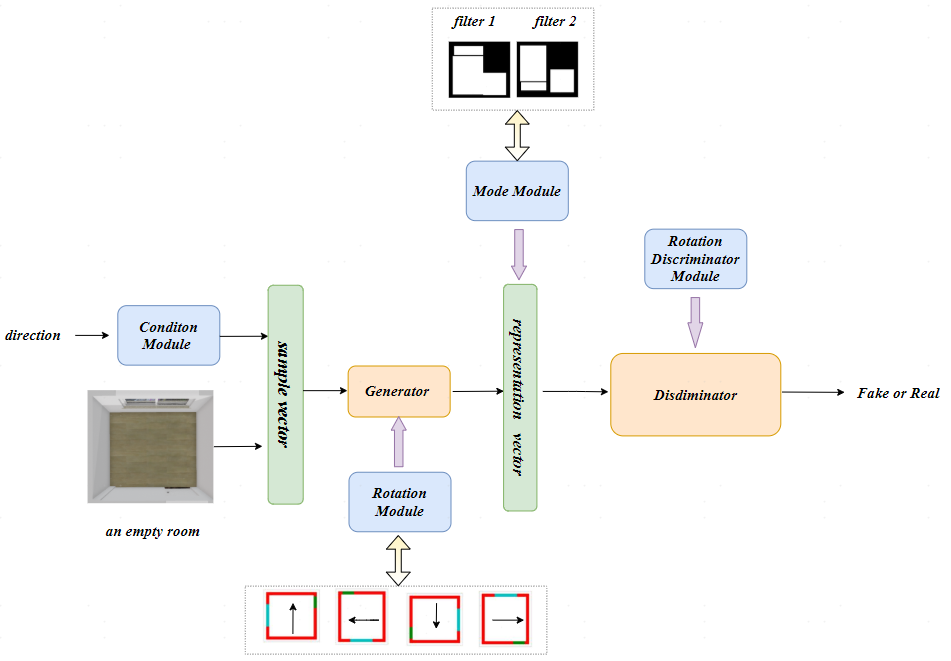}
\caption{Architecture of the proposed adversarial model which consists of a conditional adversarial module, rotation module, mode module, and a rotation discriminator module.}
\label{fig3}
\end{figure}

\section{Methods}
In this section, we propose an adversarial model to produce the layout with direction of each room. The proposed model consists of the following modules: a conditional adversarial module \cite{Lin_2018_CVPR} with a generator and a discriminator; a rotation module with several rotation filters; a mode module with two mode filters; and a rotation discriminator module. The proposed model as well as the modules are shown in Figure \ref{fig3}. In the following, we will discuss those modules as well as the training objective.

\subsection{Conditional adversarial module}
In conditional adversarial model \cite{Lin_2018_CVPR}, the generation part $g$ gets the input of a rendered image of a empty room. The condition part $v$ encodes the direction of the room as a vector, and the discriminator part $d_{1}$ determines whether the generated layout is real as illustrated Figure \ref{fig3}. 

\begin{figure}
\centering
\includegraphics[height=10cm]{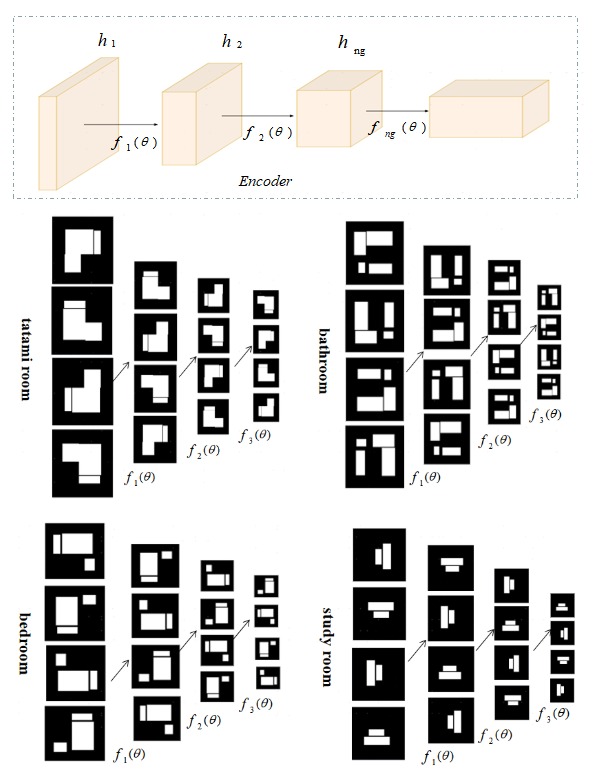}
\caption{The rotation module consists of several rotation filters. We illustrate four examples of rooms that are represented with rotation filters applied to the hidden layer.}
\label{fig4}
\end{figure}

\subsection{Rotation module}
The rotation module consists of several rotation filters. Each filter $f$ rotates the hidden representation of the generator corresponding to the rotation of a given room $\theta_{in}$ as shown in Figure \ref{fig4}. This module helps the generator to produce layout of a room with different directions, and can be written in the following form:
\begin{equation*}
    g_{rot}(\theta)=h_{n_g} \otimes f_{n_g} \dots h_{1} \otimes f_{1}(\theta)(\theta)
\end{equation*}
where $\theta$ describes the rotation of a room, $h_{1},\dots,h_{n_g}$ are $n_{g}$ hidden layers, $f_{1},\dots,f_{n_{g}}$ are $n_{g}$ rotation filters applied after each hidden layer.

\begin{figure}
\centering
\includegraphics[height=2.2cm]{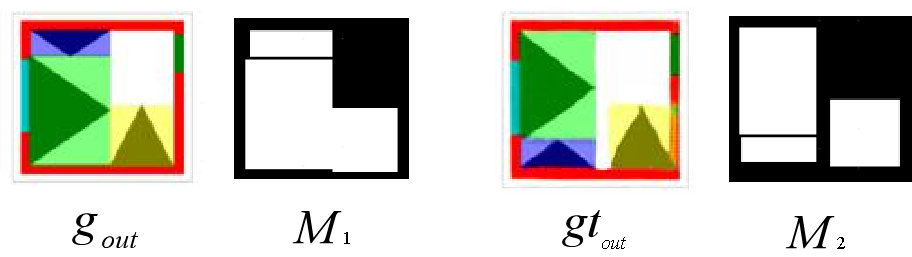}
\caption{Examples of the corresponding generated layout $g_{out}^{r}$ and the ground truth $gt^{r}$ are represented.}
\label{fig5}
\end{figure}

\subsection{Mode module}
The mode module consists of two mode filters. Each filter produces a binary attention map according to the ground truth as shown in Figure \ref{fig5}. The position inside the box of the furniture is labeled as $1$, and the left position is labeled as $0$. One filter $M_{1}$ is before the rotation discriminator. The other filter  $M_{2}$ is applied after the ground truth layout. This module helps the adversarial model to maintain the same furniture in corresponding with ground truth with rotation. The generated layout $g_{out}^{r}$ and ground truth $gt_{out}^{r}$ with these two filters can be formulated as:
\begin{equation*}
    g_{out}^{r}(\theta) = g_{out} \otimes M_{1}(\theta)
\end{equation*}
\begin{equation*}
    gt_{out}^{r}(\theta) = gt_{out} \otimes M_{2}(\theta)
\end{equation*}
where $\theta$ is the direction of the indoor scene.

\subsection{Rotation discriminator}
The rotation discriminator module adds an extra discriminator $d_{1}$ to the adversarial model. The extra discriminator determines whether the generated layout is rotated in corresponding with the same degree as the ground truth, as well as decides whether the number and category of the furniture in the layout are collapsed during rotation. 

\subsection{Training objectives}
We let $g$ denote the generator of the conditional adversarial model, $d_{1}$ denote its discriminator, and $d_{2}$ denote the rotation discriminator. We further let $F=\{f_{1},f_{2},\dots,f_{n}\}$ denote the rotation filters, $m_{1}$ denote the first mode filters, and $m_{2}$ denote the second mode filters. Given an rendered image $x_{i}$ of size $H \times W \times 3$, where $H$ and $W$ denote the height and width of the rendered image. The adversarial network model is denoted as $M(*)$. Suppose that the generator has $n_{g}$ levers, and then the generator with application of the rotation filter in each hidden layer can be formulated as $g_{rot}(\theta)=h_{1}f_{1},\dots,h_{n_g}f_{n_g}$ where $\theta$ is the rotation of a room, $h_{1},\dots,h_{n_g}$ are the $n_{g}$ hidden layers, and $f_{1},\dots,f_{n_{g}}$ are the $n_{g}$ rotation filters applied after each hidden layer. The first discriminator $d_{1}$ is applied to determine whether the generated layout image is real. Similarly, the first mode filter $M_{1}$ is applied before the generated layout $g_{out}$. It transfers the generated layout $g_{out}$ to $g_{out}^{r}$. The second mode filter $M_{2}$ is applied to the ground truth layout $gt$. It transfers the ground truth to $gt^{r}$.

\subsubsection*{Rotation discriminator network training} To train the first discriminator network $d_{1}$, the first discriminator loss $L_{1D}^{r}$ has the following form
\begin{equation*}
L_{1D}^{r} = -(1-y_{1n}^{r})\log(D(P_{n1}^{0})) + y_{1n}^{r}\log(D(P_{n1}^{1}))    
\end{equation*}
where $y_{1n}^{r}=0$ if sample $P_{n}^{0}$ is drawn from the generator, and $y_{1n}^{r}=1$ if the sample $P_{n}^{1}$ is from the ground truth. Here, $P_{n}^{0}$ denotes the rendered layout image generated from the generator with rotation $r$, and $P_{n}^{1}$ denotes the rendered ground truth layout with rotation $r$.

\subsubsection*{Mode discriminator network training} To train the second discriminator network $d_{2}$, the second discriminator loss $L_{2D}^{r}$ can be written as following:
\begin{equation*}
L_{2D}^{r} = -(1-y_{2n}^{r})\log(D(P_{n2}^{0})) + y_{2n}^{r}\log(D(P_{n2}^{1}))    
\end{equation*}
where $y_{2n}^{r}=0$ if sample $P_{n2}^{0}$ is drawn from the generator, and $y_{2n}^{r}=1$ if the sample $P_{n2}^{1}$ is from the ground truth. Here, $P_{n2}^{0}$ is drawn from the generator after the application of the filter $M_{1}$, and $P_{n2}^{1}$ is drawn from the generator after the application of the filter $M_{2}$.

\subsubsection*{Rotation generator training.} To train the rotation generator network, a conditional loss function $L_{g}^{r}$ has the following form
\begin{equation*}
L_{g}^{r} = L_{gc}^{r} + \lambda_{adv1}L_{adv1} + \lambda_{adv2}L_{adv2}
\end{equation*}
where $L_{gc}^{r}$ and $L_{adv}$ denote the generation loss with rotation and the adversarial loss, respectively. Here $\lambda_{adv1}$ and $\lambda_{adv2}$ are two constants for balancing the multi-task training.

Given the rendered indoor scene $x_{i}$ and its rotation $\theta_{i}$, ground truth $y_{i}$ and prediction results $g_{outi}$, the generator loss is 
\begin{equation*}
L_{gc}^{r} = -y_{i} \log(g_{outi})
\end{equation*}
Moreover, the $L_{adv1}$ and $L_{adv2}$ can be written as
\begin{equation*}
L_{adv1} = -\log(D(P_{n1}^{1}))
\end{equation*}
\begin{equation*}
L_{adv1} = -\log(D(P_{n2}^{1}))
\end{equation*}
During the training, the adversarial loss is used to fool the discriminator by maximizing the probability of the generated prediction being considered as the ground truth distribution.

\section{Proposed dataset}
\setcounter{footnote}{0}
In this paper, we propose a dataset of indoor furniture layouts together with an end-to-end rendering image of the interior layout\footnote{The dataset and  codes will be released soon.}. These layout data is from designers at the real selling end where proprietors choose the design of the layout for their properties.

\begin{figure}
\centering
\includegraphics[height=9cm]{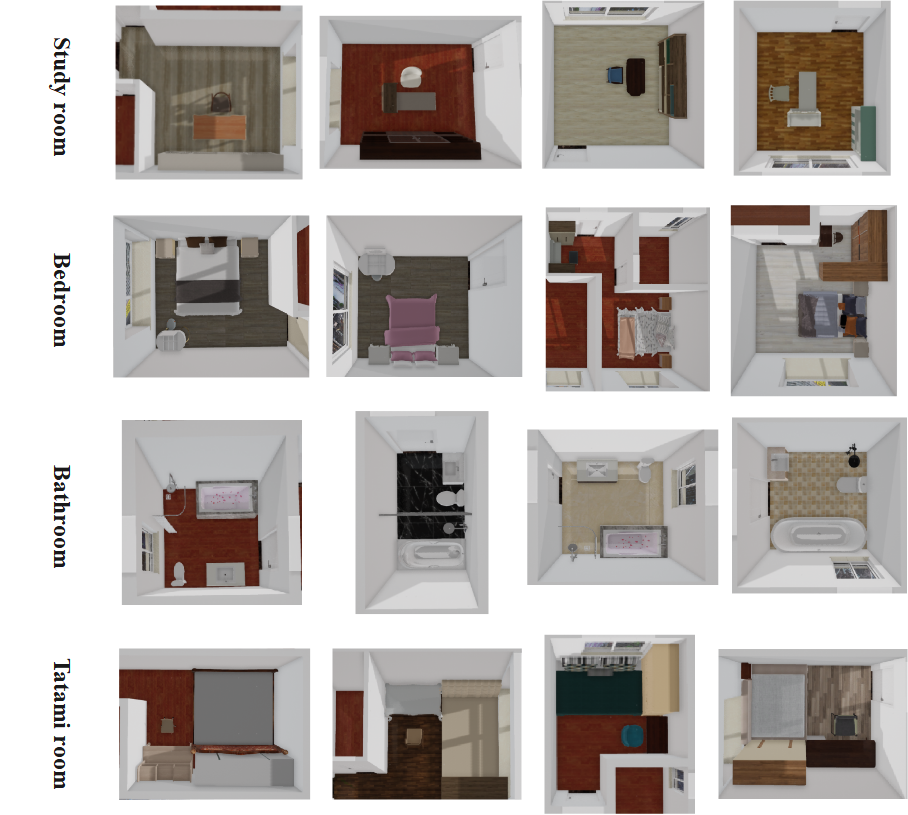}
\caption{Samples from the proposed indoor-layout dataset including four types of rooms: the study room, the bedroom, the tatami room and the bathroom. The rooms are oriented to different directions.}
\label{fig6}
\end{figure}

\subsection{Interior Layouts}
$1356$ professional designers work with an industry-lever virtual tool to produce a variety of designs. Among these designs, a part of them are sold to the proprietors for their interior decorations. We collect these designs at the selling end and provide $4800$ interior layouts. Each sample of the layout has the following representation including the categories of the furniture in a room, the position $(x,y)$ of each furniture, the direction $(rot)$ of each furniture, the position $(x,y)$ of the doors and windows in the room, and the position $(x,y)$ of each fragment of the walls. Figure \ref{fig6} illustrates the samples of layouts adopted from the interior design industry and sold to the proprietors. It contains $4$ types of rooms including the bedroom, the bathroom, the study room and the tatami room. The designs of these rooms are sold to the proprietors whose properties have $2-4$ rooms and $2-3$ bathrooms. Besides, each designs are modified after several versions both following the professional designers knowledge and the personalized suggestions of the each proprietor. 

\begin{figure}
\centering
\includegraphics[height=7.5cm]{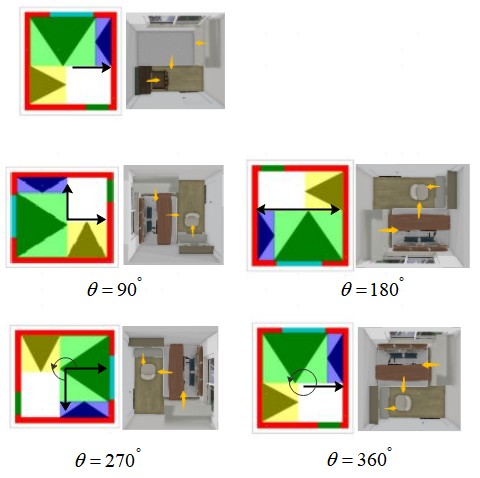}
\caption{Samples from the indoor-layout dataset are rotated in four directions $\theta= \ang{90}, \ang{180}, \ang{270}, \ang{360}$.}
\label{fig7}
\end{figure}

Besides, all $4800$ designs are rotated in $4$ directions for $ \theta=\ang{90}, \ang{180}, \ang{270}$ and $\ang{360} $. The position $(x,y)$ of each furniture, the direction $(rot)$ of each furniture, the position $(x,y)$ of the doors and windows in the room, and the position $(x,y)$ of each fragment of the walls are all rotated, resulting in a total number of the layouts $14400$. Some examples are shown in Figure \ref{fig7}. 
\begin{figure}
\centering
\includegraphics[height=7.5cm]{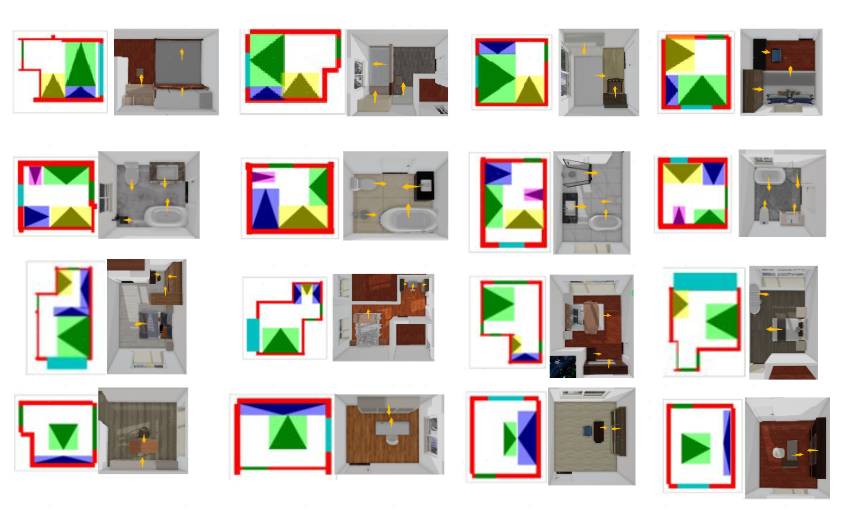}
\caption{The layout samples and the corresponding rendered scenes are represented. For each sample, the layout sample including the position, the direction, the size of each furniture are represented on the left side, the corresponding rendered scene is represented on the right side.}
\label{fig8}
\end{figure}

\subsection{Rendered Layouts}
Each layout sample is corresponding to the rendered layout images. These images are the key demonstration of the interior decoration. These rendered images contain several views and we collect the top-down view as the rendered view as shown in Figure \ref{fig8}. Therefore, the dataset also contains $14400$ rendered layouts in the top-down view. Each rendered layout is corresponding to a design. The rendered data is produced from an industry-lever virtual tool which has already provided missions of rendering layout solutions to the proprietors.

\section{Evaluation}
In this section, we present qualitative and quantitative results demonstrating the utility of our proposed adversarial model for scene synthesis. Four types of indoor rooms are evaluated including the bedroom, the bathroom, the study room and the tatami room. $4000$ samples are randomly chosen for training, and $800$ samples are used for the test. Both the training and test rooms are rotated in $4$ directions: $\ang{90}$, $\ang{180}$, $\ang{270}$ and $\ang{360}$. The first baseline model is a classical adversarial model \cite{isola2017image} which takes a pair of samples of a rendered empty room and its layouts for training. For the inference, it produces the layout of furniture given the rendered empty room. The second baseline model is a conditional adversarial model \cite{Lin_2018_CVPR}, which takes the pair of samples together with the rotation $\theta$ for training. For the inference, it encodes the direction of the room $\theta$ and the rendered empty room $x_{i}$ and produces the layout. Similarly, our model encodes $\theta$ and $x_{i}$ to produce the layout.

\subsection{Evaluation metrics}
For the task of interior scene synthesis, we apply three metrics for the evaluation. Firstly, we use average mode accuracy for the evaluation. It is to measure the accuracy of category of furniture for a layout in corresponding with the ground truth. This average mode accuracy is defined as
\begin{equation*}
\text{Mode} = \frac{\sum_{i=1}^{n}N_{i}^{1}}{\sum_{i=1}^{n}N_{i}^{total}} 
\end{equation*}
where $N_{i}^{total}$ is the total number of $ith$ category of furniture in the ground truth dataset, and $N_{i}^{1}$ is the number of the $ith$ category of furniture in the generated layout in corresponding with the ground truth. For example, if the $ith$ furniture is in the predicted layout where the ground truth layout also contains this furniture, then it is calculated. Note that $n$ is the total number of the category of the furniture.

Secondly, in order to evaluate the position accuracy of furniture layout, we apply the classical mean Average Precision (mAP) to measure the position of the furniture in the predicted layout. Note that the threshold for the Intersection over Union (IoU) between the predicted box of $ith$ furniture and the ground truth box is set to $0.5$.

Thirdly, we define a metric called RoT to measure the rotation accuracy of each furniture in the prediction. At the industry end, the direction of the furniture is also a key for the interior designs. For example, the TV set should be placed towards inside the room. The RoT is defined as
\begin{equation*}
\text{RoT}= 1 - \frac{\sum_{j=1}^{n_{total}}|rot(pred_{j})-rot(gt_{j})|}{n_{total}\times90 }
\end{equation*}
where $n_{total}$ is the total number of furniture in the dataset, $rot(pred_{j})$ is the rotation of the $jth$ furniture in the prediction, and $rot(gt_{j})$ is the rotation of the corresponding furniture in the ground truth.

\begin{figure}
\centering
\includegraphics[height=5.4cm]{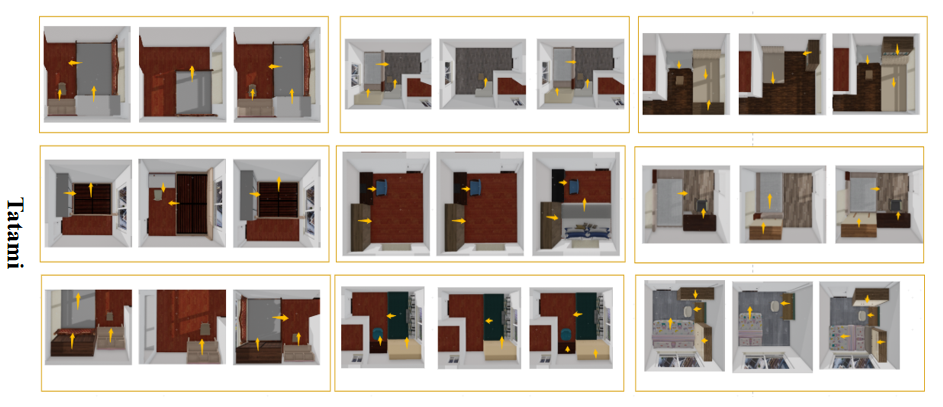}
\caption{Layouts produced by our proposed model and two baselines for tatami. For each comparison sample, the left layout is from the baseline 1 model, the middle layout is from the baseline 2 model, and the right layout is from our proposed model.}
\label{fig9_1}
\end{figure}

\begin{figure}
\centering
\includegraphics[height=5.3cm]{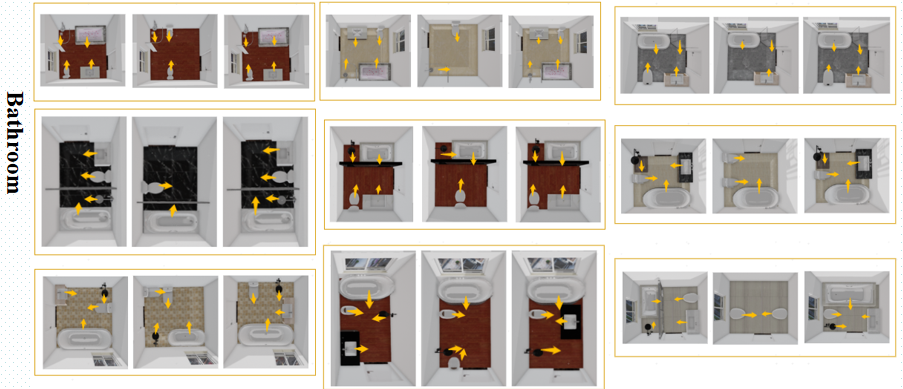}
\caption{Layouts produced by our proposed model and two baselines for bathroom. For each comparison sample, the left layout is from the baseline 1 model, the middle layout is from the baseline 2 model, and the right layout is from our proposed model.}
\label{fig9_2}
\end{figure}

\begin{figure}
\centering
\includegraphics[height=6.8cm]{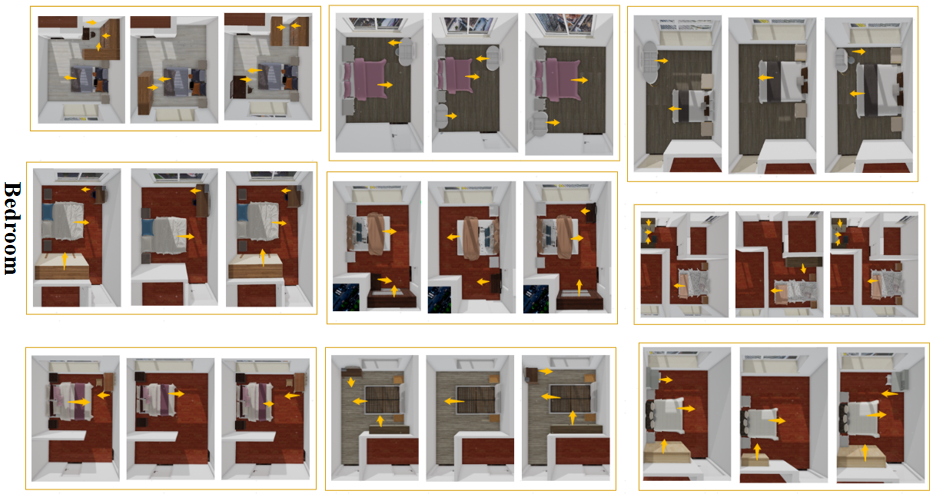}
\caption{Layouts produced by our proposed model and two baselines for bedroom. For each comparison sample, the left layout is from the baseline 1 model, the middle layout is from the baseline 2 model, and the right layout is from our proposed model.}
\label{fig9_3}
\end{figure}

\begin{figure}
\centering
\includegraphics[height=2.0cm]{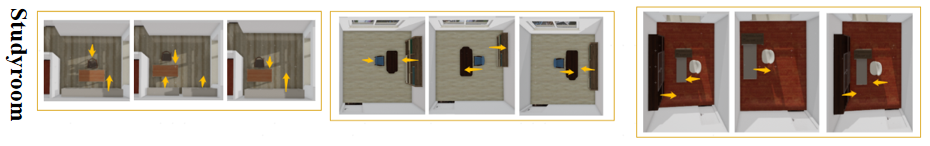}
\caption{Layouts produced by our proposed model and two baselines for study room. For each comparison sample, the left layout is from the baseline 1 model, the middle layout is from the baseline 2 model, and the right layout is from our proposed model.}
\label{fig9_4}
\end{figure}

\begin{table}
\centering
\begin{tabular}{|p{2cm}|p{1cm}|p{1cm}|p{1cm}|p{1cm}|p{1cm}|p{1cm}|p{1cm}|p{1cm}|p{1cm}|}
 \hline
 \multicolumn{1}{|c|}{}&\multicolumn{3}{|c|}{\text{Mode}}&\multicolumn{3}{|c|}{\text{mAP}}&\multicolumn{3}{|c|}{\text{RoT}}\\
 \hline
 \hfil model    &\hfil base1  & \hfil base2  &\hfil ours   &\hfil base1 &\hfil base2 &\hfil ours  &\hfil base1  &\hfil base2  &\hfil ours  \\
 \hline
 \hfil tatami   &0.7862 &0.9326 &0.9565 &0.626 &0.625 &0.726 &0.5860 &0.6913 &0.7613\\
 \hfil bathroom &0.7522 &0.8545 &0.8645 &0.506 &0.538 &0.708 &0.4563 &0.7020 &0.7861\\
 \hfil bedroom  &0.7563 &0.7242 &0.8871 &0.585 &0.527 &0.782 &0.4287 &0.6826 &0.7864\\
 \hfil study    &0.7444 &0.8885 &0.9000 &0.472 &0.575 &0.775 &0.4419 &0.6625 &0.7704\\
 \hline
\end{tabular}
\label{table1}
\caption{Evaluation metrics for different methods on four types of rooms.}
\end{table}

We compare with two baseline models for scene synthesis for four types of rooms. The results are shown in Figure \ref{fig9_1} -- \ref{fig9_4}. Our model outperforms the baseline models in the following aspects. Firstly, for rotated indoor room, our model predicts the same category of the furniture with the ground truth layout, while the two baseline models lose the category of the furniture. Secondly, our model predicts a good position of each furniture during the rotation of the room, while the baseline models sometimes predicts unsatisfied position that is strongly against the knowledge of the professional interior designers. Thirdly, the baseline models sometimes fail to produce the position and the size of furniture while our model seldom yield this failure.

We also compare with two baseline models quantitatively. All three performance metrics for four types of room are given in Tables \ref{table1}, which shows the accuracy of mode, position and size, and the direction of furniture in the predicted layout. Our model outperforms the baseline models in all metrics for all types of rooms.  

\section{Discussion}
In this paper, we presented an adversarial model to predict the interior scene synthesis with rotation. In addition, we propose an interior layouts dataset that all the designs are drawn from the professional designers. The proposed model achieves the best performance among baselines on the interior layouts dataset. There are several avenues for the future work. Our method is currently limited to the generation of layouts for the common rooms, and the layout of other rooms is hard to predict. For example, it is difficult to predict the layout for the luxury bedroom where the bathroom and the cloakroom are also built in the luxury bedroom. Besides, the furniture category for each type of the room is limited. It is worthwhile to extend our work and study a more general setting where more furniture such as dressing table, office desk or a leisure sofa are included.

\bibliographystyle{splncs04}
\bibliography{egbib}
\end{document}